# Use of Metamorphic Relations as Knowledge Carriers to Train Deep Neural Networks

Tsong Yueh Chen [a], Pak-Lok Poon [b], Kun Qiu [c], Zheng Zheng [d], Jinyi Zhou [d]

[a] *Department of Computer Science & Software Engineering, Swinburne University of Technology, Australia*
[b] *School of Engineering & Technology, Central Queensland University, Melbourne, Australia*
[c] *School of Electrical Engineering & Automation, Hefei University of Technology, China*
[d] *School of Automation Science & Electrical Engineering, Beihang University, China*

**ABSTRACT**

Training multiple-layered deep neural networks (DNNs) is difficult. The standard practice of using a large number of samples for training often does not improve the performance of a DNN to a satisfactory level. Thus, a systematic training approach is needed. To address this need, we introduce an innovative approach of using metamorphic relations (MRs) as "knowledge carriers" to train DNNs. Based on the concept of metamorphic testing and MRs (which play the role of a test oracle in software testing), we make use of the notion of metamorphic group of inputs as concrete instances of MRs (which are abstractions of knowledge) to train a DNN in a systematic and effective manner. To verify the viability of our training approach, we have conducted a preliminary experiment to compare the performance of two DNNs: one trained with MRs and the other trained without MRs. We found that the DNN trained with MRs has delivered a better performance, thereby confirming that our approach of using MRs as knowledge carriers to train DNNs is promising. More work and studies, however, are needed to solidify and leverage this approach to generate widespread impact on effective DNN training.

## 1. Introduction

Training multi-layered deep neural networks (DNNs) is well known to be difficult [9]. Various researchers in the field have proposed their own training approaches to address this issue [13]. Despite such effort, DNNs are still reported to produce incorrect outputs or mistakes, largely because of the absence of a systematic and thorough training approach. It is a standard practice to train a DNN with a massive amount of training samples, so that a trained DNN will deliver a good output performance. This standard practice has an implicit assumption—a massive training set is "expected" to cover a large variety of input scenarios, such that a trained DNN can produce a correct output for any input. However, this assumption does not always hold. This is particularly the case when a DNN has many hidden layers, because it is very difficult to find training samples which can penetrate through to the deep layers of a DNN. Empirical studies [13, 22] have found that poor outputs will be produced from DNNs even with three or more hidden layers. Furthermore, training a DNN with a massive training set will incur a significant amount of training time, which may not be affordable in some real-world situations.

In view of the above problems, we propose to use metamorphic relations (MRs) as "knowledge carriers" to train a DNN system. In this paper, we will restrict our discussions to DNN classifier systems, which will be simply referred to as "DNNs". Our preliminary experiment has shown that a DNN trained with MRs has a better performance (that is, less misclassifications) than another DNN trained without MRs.

## 2. Metamorphic testing and metamorphic relation

Metamorphic testing (MT) has been successfully used to *test* a software system with the *test oracle* problem (the situation where the expected outcome of an input is infeasible to be determined) [4, 20]. Since its invention two decades ago, MT has been increasingly accepted as a mainstream testing technique and widely used in different application domains and platforms, including web services [1, 17], biomedical applications [3], embedded systems

*E-mail addresses:* tychen@swin.edu.au **(T.Y. Chen)**, p.poon@cqu.edu.au **(P.-L. Poon)**, qiukun@hfut.edu.cn **(K. Qiu)**, zhengz@buaa.edu.cn **(Z. Zheng)**, zy1803193@buaa.edu.cn **(J. Zhou)**.

[12], security [5, 18], compilers [14, 16], component-based software [2], machine learning systems [23, 25], online search engines [30], computer vision [26], and autonomous car systems [21, 27, 28]. More recently, MT has also gone beyond testing, and has been extended to other disciplines such as program proving [6, 8], fault localization [24], program repair [10, 11], and program understanding [29].

In essence, MT was first developed as a software testing technique which involves two or more test inputs, thereby involving multiple program executions. A core task of MT is the identification of metamorphic relations (MRs) [19]. Each MR is a property of the program under test ($P$) which explicitly states the relationship between the *source inputs*, their associated follow-up inputs, as well as their corresponding outputs (*source outputs* and *follow-up outputs*). For simplicity, in this paper, each MR is assumed to have only one source input ($s$) and one follow-up input ($f$). With respect to an MR, $P$ is executed twice: firstly with $s$ and then with $f$. If the source output corresponding to $s$ and the follow-up output corresponding to $f$ do not comply with MR, we can conclude that $P$ is faulty.

Consider, for example, an implementation *sin* of the sine function. Suppose we execute *sin* with an input of 28° and obtain an output of 0.4695 (rounded to four decimal places). It is difficult to determine whether or not this output (i.e., $sin(28°) = 0.4695$) is correct, because the "precise" sine value of an angle is normally unknown unless it is a special angle such as 90° where $sin(90°) = 1$. However, there exists a property of sine that $sin(x) = sin(x + 360°)$, where $x$ is any angle in degree measure. From this property, an $MR_a$ is defined: "If $y = x + 360°$, then $sin(x) = sin(y)$." Here, $x$, $y$, $sin(x)$, and $sin(y)$ are called the *source input*, the *follow-up input*, the *source output*, and the *follow-up output*, respectively. In view of $MR_a$, we execute *sin* twice: firstly with any angle $x$ (e.g., 28°) as a source input; and then with the angle $y$ (e.g., 388° = 28° + 360°) as a follow-up input. If the two outputs are not equal (i.e., $sin(28°) \neq sin(388°)$), then we know that *sin* is faulty, even though we do not know whether or not $sin(28°)$ (or $sin(388°)$) should be equal to 0.4695. For each MR, its source inputs and corresponding follow-up inputs form a *metamorphic group of inputs* (or simply a *metamorphic group* and is denoted by $MG$). In the above example, (28°, 388°) is an MG for $MR_a$. Obviously, for a given MR, it often has many MGs. Also, it is not difficult to see that $MR_a$ carries a piece of knowledge, that is, a periodic property of sine that can be used for testing *sin*.

## 3.  Related Work

Besides using a massive training set, various studies (e.g., [7]) have been proposed to improve the effectiveness of DNN training by means of *adversarial examples* (those samples which are slightly perturbed from either the existing training samples or the testing samples, and will result in some abnormal behaviors of the DNN). After these adversarial examples have been identified, they are used as additional training samples for a DNN.

Since adversarial samples are effectively failure-causing follow-up test cases (in the context of MT) and have been used as additional training samples for DNNs, researchers are motivated to investigate how to use follow-up test cases as additional training samples for DNNs (e.g., [7, 21, 27]). However, such addition not only increases the "size" of the training set, but also increases the "variety" of the training samples. Intuitively speaking, increasing the size and variety of the training set will make this set more comprehensive, thereby causing the training more robust. Indeed, study results have shown that this training approach improves the reliability and robustness of DNNs.

## 4.  Our approach of using MRs as knowledge carriers to train DNNs

MT was first proposed as a method to select test cases and to alleviate the test oracle problem in software testing [4]. Since its inception, the application of MT has gone beyond software testing to other disciplines such as program proving [6, 8], fault localization [24], program repair [10, 11], and program understanding [29]. Along with the evolution of MT application to other areas, the role of MRs has also evolved from being a test oracle in the context of software testing to other appropriate concepts in the relevant areas. For example: (a) in validation, MRs play the role of the expectations from the users [25]; (b) in system understanding, MRs play the role of a program specification [29]; and (c) in program selection, MRs play the role of the evaluation's adequacy criteria [25, 30]. Inspired by the evolving roles of MRs across different disciplines, we conceive that MRs can play the role of knowledge carriers in the context of DNN training.

Consider, for example, a DNN (denoted by AML) which takes in an image of any animal (e.g., a lion) and attempts to determine its species. Suppose we have an image (denoted by $s$) of an animal species as the source image (or input). An MR (denoted by $MR_b$) can be defined for AML: "Given a source image $s$, if we generate its mirror image as a follow-up image $f$, then the source output label (or simply the source output) and the follow-up output label (or simply the follow-up output) generated from AML should be the same". For training AML, with $MR_b$, we construct two training samples: the first sample includes the source image $s$ added with its corresponding source output (e.g., the label "lion"); the second sample includes the follow-up image $f$ added with its corresponding follow-up output (e.g., the label "lion"). Intuitively speaking, the training samples so constructed are more effective than any two randomly generated images (denoted by $r_1$ and $r_2$) for training AML. This is because, not only each of $s$ and $f$ represents an "individual" piece of training knowledge (as with $r_1$ and $r_2$), the *relationship* between $s$, $f$, and their corresponding outputs as identified in this case carries *additional* knowledge, that is, the original and mirror images should refer to the same object in the above example. Such additional knowledge (or constraint or property) is then intrinsically fed into AML. This is the intuition which leads to our hypothesis that MRs can serve as knowledge carriers to effectively train a DNN (or more specifically, as a new way to feed the logic to a DNN).

In our approach, the ultimate objective is to train a DNN with MRs so that it has better performance. Since an MR is an abstraction of knowledge (as can be seen in $MR_a$ and $MR_b$ above), therefore it cannot be directly used for DNN training. To alleviate this problem, we generate a set of training samples from the MGs associated with an MR, because MGs are instances of an MR. We use the components of an MG, that is, the source image and the follow-up image, as well as their corresponding (output) labels to form the samples for training. More specifically, with reference to an MG, we construct two training samples: the first sample consisting of the source image and its corresponding label; the second sample consisting of the follow-up image and its corresponding label. In this way, the knowledge attached to an MR can be passed through the components of its MG to a DNN during training. This is the first paper which proposes to use MRs (through their associated MGs) as knowledge carriers to train DNNs (in particular, their logic). Note that our approach fundamentally differs from previous related studies (e.g., [7, 21, 27]) which focused on the use of the follow-up test cases, in particular, the failure-causing follow-up test cases for training DNNs. Our approach makes use of the whole MGs to capture the complete knowledge of MRs, rather than the piecemeal knowledge associated with the individual follow-up inputs. In other words, the previous studies used a "data-oriented" approach towards training of DNNs, which differs from our "property-oriented" approach discussed in this paper. The contributions of this study are: (a) the approach of using MRs as knowledge carriers to train DNNs, and (b) the experimental investigation to validate the hypothesis that such a training approach is effective.

## 5. Our preliminary experiment

We performed a preliminary experiment to verify our hypothesis that a DNN trained with MRs has better performance than another DNN trained without MRs. Below we outline the experimental setup:

a. We used a large set ($S$) of $k$ training samples (denoted by $t_1$, $t_2$, ..., $t_k$).
b. We generated another set ($S'$) of $k$ training samples from $S$ using the following steps:
   i. $m$ (where $m$ is an even number and $m<k$) training samples were randomly removed from $S$. After removal, the resulting set $S_1 = S \setminus \{t_1, t_2, ..., t_m\}$. The size of $S_1$ was reduced to ($k - m$).
   ii. We randomly deleted half of the samples from the set $\{t_1, t_2, ..., t_m\}$ to form a smaller set $S_2$ (with a size of $m/2$).
   iii. We defined an MR for the DNN. For each sample in $S_2$, we considered it as a source image and then generated a follow-up image with respect to MR. All the generated follow-up images were added back to $S_2$. Note that, after the addition, the size of $S_2$ was $m$.
   iv. We generated a set ($S'$) of training samples such that $S' = S_1 \cup S_2$. Note that both $S_1$ and $S_2$ contained $k$ training samples.
c. We trained a version of DNN (denoted by $DNN_1$) with $S$. We then trained another version of DNN (denoted by $DNN_2$) with $S'$.
d. We used a new sample set $S''$ ($\neq S \neq S'$) to test the output performance of $DNN_1$ and $DNN_2$.

LeNet5 [15] (which is a convolutional neural network) was chosen as the DNN in our experiment. We used MNIST [15] as our data set, which contained 70,000 gray images of handwritten digits. This data set was split into two pools: the first pool containing 60,000 images from which $S$ (in step a above) was randomly selected for DNN training, and the second pool containing 10,000 images from which $S''$ (in step d above) was randomly selected for evaluating the performance of a DNN after training. Five MRs were defined and used in the preliminary experiment. These MRs involved rotating, shifting, scaling, vertical mirroring, and elastic distorting, respectively, to construct the follow-up images from the source images. The transformations corresponding to the first four MRs are obvious from their names and, hence, need no further explanation. For the elastic-distorting MR, it simulated image distortion caused by shaking hands when writing. When applying the vertical-mirroring MR to transform numeric digits, only those digits which are vertically symmetrical (i.e., 0, 1, and 8) are applicable.

A one-tailed t-test was used to compare and analyze the difference of the accuracy performance between $DNN_1$ (trained without using MRs) and $DNN_2$ (trained with MRs). When $DNN_2$ had been separately trained with the rotating MR, the shifting MR, the scaling MR, the vertical-mirroring MR, and the elastic-distorting MR, following by comparing the performance between DNN$_2$ and DNN$_1$, the corresponding *p*-values were 0.000 (when $k = 1{,}500$ and $\frac{m}{2k} = 6\%$), 0.009 (when $k = 15{,}000$ and $\frac{m}{2k} = 3\%$), 0.006 (when $k = 1{,}500$ and $\frac{m}{2k} = 3\%$), 0.042 (when $k = 1{,}500$ and $\frac{m}{2k} = 12\%$), and 0.002 ($k = 1{,}500$ and $\frac{m}{2k} = 25\%$), respectively. (Note that the value of $\frac{m}{2k}$ denotes the ratio of the number of follow-up samples to the number of training samples.) This indicates that DNN$_2$ outperformed DNN$_1$, and the difference in performance was statistically significant at a 0.05 level. Note that the comparison of performance between DNN$_1$ and DNN$_2$ was fair because both of them were trained with the same number of samples.

## 6. Summary and Conclusion

The main contribution of this paper is to introduce an innovative application of MT. More specifically, we discussed and explained how to use MRs (through their associated MGs) to feed useful knowledge and logic to DNNs during training. As such, MRs play the role of knowledge carriers for DNN training. A preliminary experiment has shown that our training approach is promising and under certain settings, the DNN so trained has a better performance when compared with another DNN trained without using MRs.

Our "MR-based" (and, hence, "property-oriented") training approach has opened up a new line of research or avenue for effective DNN training, by solving the problem of feeding knowledge (or logic) to multiple layers of a DNN. It would be worthwhile to further explore the contributions of MTs to DNN training in more details, by means of more and substantial studies and experiments.

**Acknowledgments**

This work was supported in part by a 2020 Facebook Research Grant, the Australian Research Council (DP210102447), the National Natural Science Foundation of China (Grant Nos. 61772055 & 61872169), the Technical Foundation Project of Ministry of Industry & Information Technology of China (Grant No. JSZL2016601B003), and the Equipment Preliminary R&D Project of China (Grant No. 41402020102).